\documentclass[10pt,twocolumn,letterpaper]{article}

\usepackage{times}
\usepackage{geometry}
\usepackage{makecell,amsmath,amssymb,amsfonts,adjustbox,booktabs,multirow,algorithm,algorithmic,tabularx}
\usepackage[table]{xcolor}

\definecolor{compblue}{rgb}{0.21,0.49,0.74}
\usepackage[pagebackref,breaklinks,colorlinks,allcolors=compblue]{hyperref}

\title{\textbf{Beyond Artifacts: Real-Centric Envelope Modeling for Reliable \\AI-Generated Image Detection}}

\title{\textbf{Beyond Artifacts: Real-Centric Envelope Modeling for Reliable \\AI-Generated Image Detection}}

\author{
    {Ruiqi Liu}$^{1,2}$\quad 
    {Yi Han}$^{3}$ \quad
    {Zhengbo Zhang}$^{1}$ \quad
    {Liwei Yao}$^{4}$ \quad
    {Zhiyuan Yan}$^{5}$ \quad
    {Jialiang Shen}$^{6}$ \\ % 这里可以用 \\ 手动控制作者换行位置
    {ZhiJin Chen}$^{1}$ \quad
    {Boyi Sun}$^{1}$ \quad
    {Lubin Weng}$^{1}$ \quad
    {Jing Dong}$^{1}$ \quad
    {Yan Wang}$^{7}$\thanks{Corresponding author: \texttt{wangyan@air.tsinghua.edu.cn}} \quad
    {Shu Wu}$^{1}$\thanks{Corresponding author: \texttt{shu.wu@nlpr.ia.ac.cn}}
    \vspace{0.5em} \\ % 作者和单位之间的间距
    \small
    $^{1}$Institute of Automation, Chinese Academy of Sciences \quad
    $^{2}$School of Advanced Interdisciplinary Sciences, UCAS \\
    \small$^{3}$Southwest University \quad
    $^{4}$Shanghai Second Polytechnic University \quad
    $^{5}$Peking University \\
    \small$^{6}$The University of Sydney \quad
    $^{7}$Tsinghua University
}

\date{}

\begin{document}
\maketitle

\begin{abstract}
The rapid progress of generative models has intensified the need for reliable and robust detection under real-world conditions. 
However, existing detectors often overfit to generator-specific artifacts and remain highly sensitive to real-world degradations. 
As generative architectures evolve and images undergo multi-round cross-platform sharing and post-processing (chain degradations), these artifact cues become obsolete and harder to detect. 
To address this, we propose \textbf{Real-centric Envelope Modeling} (REM), a new paradigm that shifts detection from learning generator artifacts to modeling the robust distribution of real images. 
REM introduces feature-level perturbations in self-reconstruction to generate near-real samples, and employs an envelope estimator with cross-domain consistency to learn a boundary enclosing the real image manifold. 
We further build \textbf{\textit{RealChain}}, a comprehensive benchmark covering both open-source and commercial generators with simulated real-world degradation. 
Across eight benchmark evaluations, REM achieves an average improvement of 7.5\% over state-of-the-art methods, and notably maintains exceptional generalization on the severely degraded RealChain benchmark, establishing a solid foundation for synthetic image detection under real-world conditions. 
\textbf{The code and the RealChain benchmark will be made publicly available upon acceptance of the paper.}
\end{abstract}

% Extensive experiments show that REM achieves superior generalization across severe degradations and unseen generators, establishing a strong foundation for real-world synthetic image detection.    

\section{Introduction}
\label{sec:intro}

The rapid progress of generative models has greatly enhanced the realism of synthesized images, driving their widespread use in digital content creation \cite{nichol2021glide,betker2023dalle3,midjourney2023,firefly2023}.
However, this also raises increasing concerns about visual content authenticity \cite{goodfellow2014generative,ho2020denoising,yan2024df40}.
Thus, reliably detecting AI-generated images (AIGI) under real-world conditions has become an urgent challenge \cite{cozzolino2024raising,cavia2024real,yan2024orthogonal,yan2024sanity,dell2025truefake}.

\begin{figure}[t!]
\centering
\includegraphics[scale=0.20]{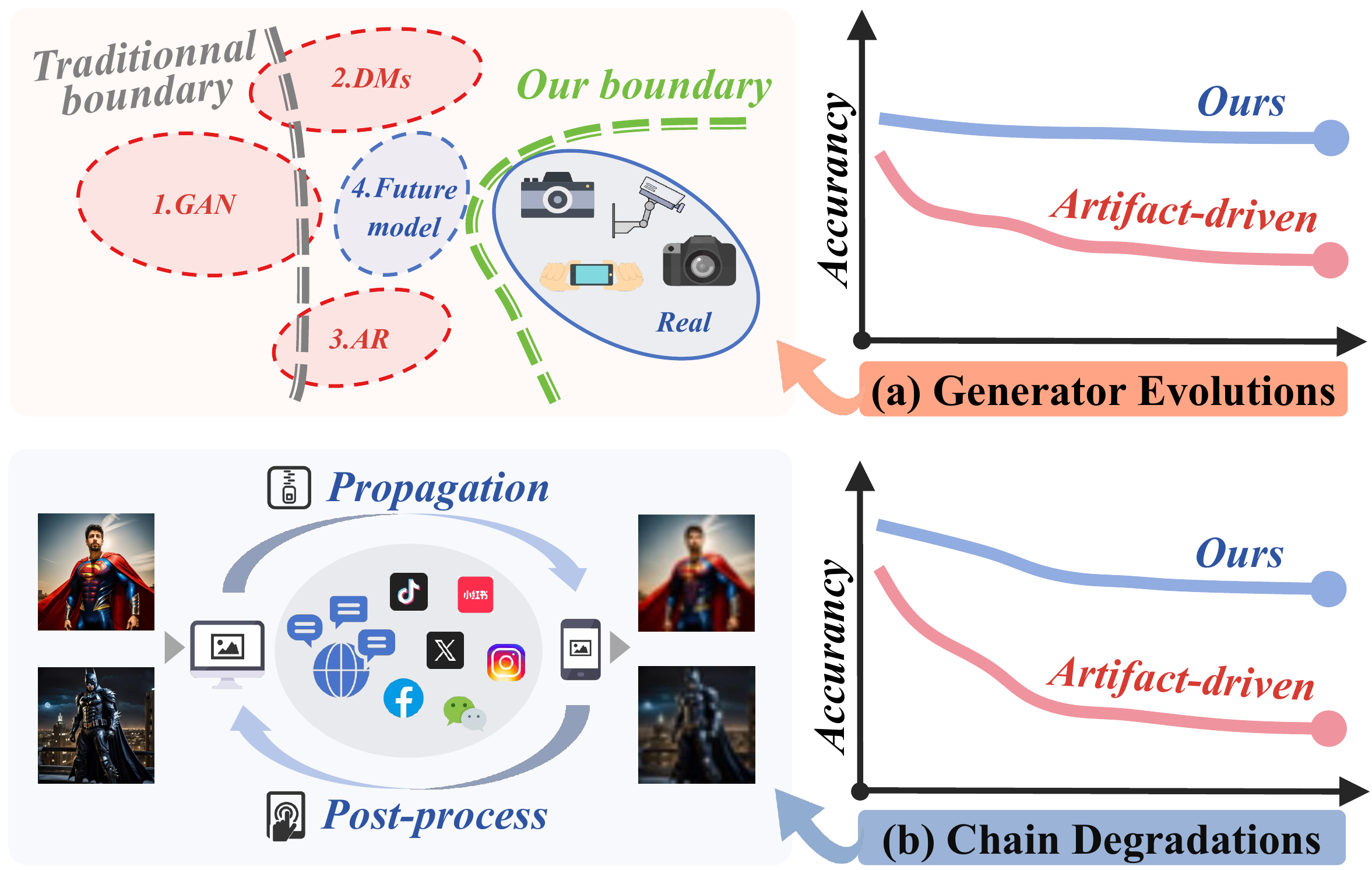}
\caption{
\textbf{Challenges of AIGI detection in real-world scenarios.}
(a) As generative models advance (from GAN models to Diffusion models to Autoregressive models), their outputs increasingly approach the real image manifold, rendering decision boundaries learned from older generators insufficient to distinguish emerging ones.
(b) Images are often compressed (e.g., JPEG) during multi-round cross-platform/device propagation and post-processed by users (e.g., filters, digital stickers). These chain degradations severely impair the performance of artifact-based detectors.
}

\label{fig_1} 

\end{figure}

Existing detectors follow a generative artifact–driven paradigm \cite{durall2020watch, frank2020leveraging, tan2024rethinking, ju2022fusing, zheng2024breaking, yang2025d, park2025community}.
These detectors simplify the synthetic image space into a finite set, attempting to learn general semantic or pixel-level forgery cues (\textbf{artifacts}) from older generators, while assuming that all images maintain uniform quality during both training and inference.
However, as shown in Fig.~\ref{fig_1}, their effectiveness drops sharply under real-world conditions, mainly due to two limitations:
(i) they tend to overfit to generator-specific patterns and struggle to generalize to unseen forgery types; and
(ii) their reliance on subtle fake cues makes them vulnerable to real-world perturbations.

\begin{figure}[t]
\centering
\includegraphics[scale=0.134]{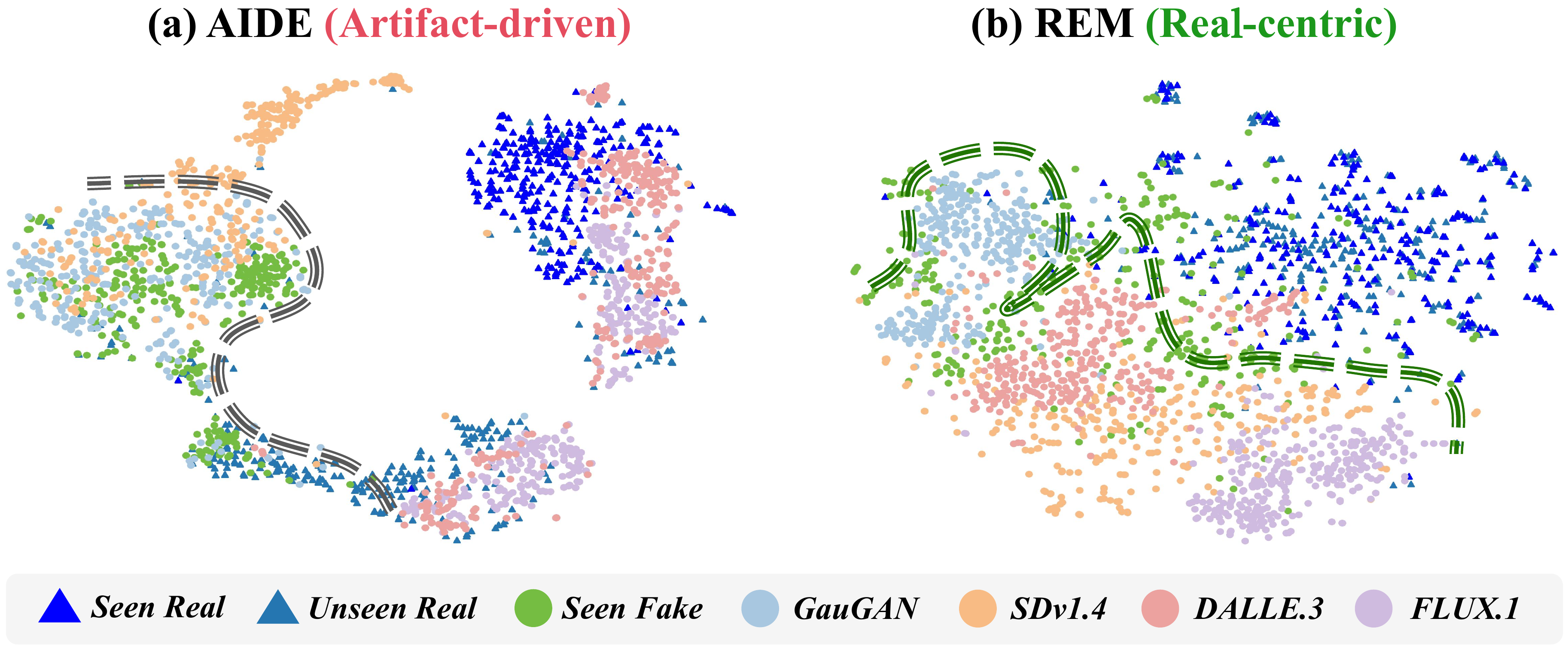}
\caption{
\textbf{Challenges of Generator Evolutions.}
New generators continually emerge with evolving architectures and sampling strategies, 
causing the feature discrepancy between real and fake images to diminish over time. 
Detectors trained on obsolete generators overfit to seen artifacts and fail to generalize to new ones, highlighting the need for a real-centric modeling paradigm.
}
\label{fig_clg2} 
\end{figure}

\begin{figure}[t!]
\centering
\includegraphics[scale=0.13]{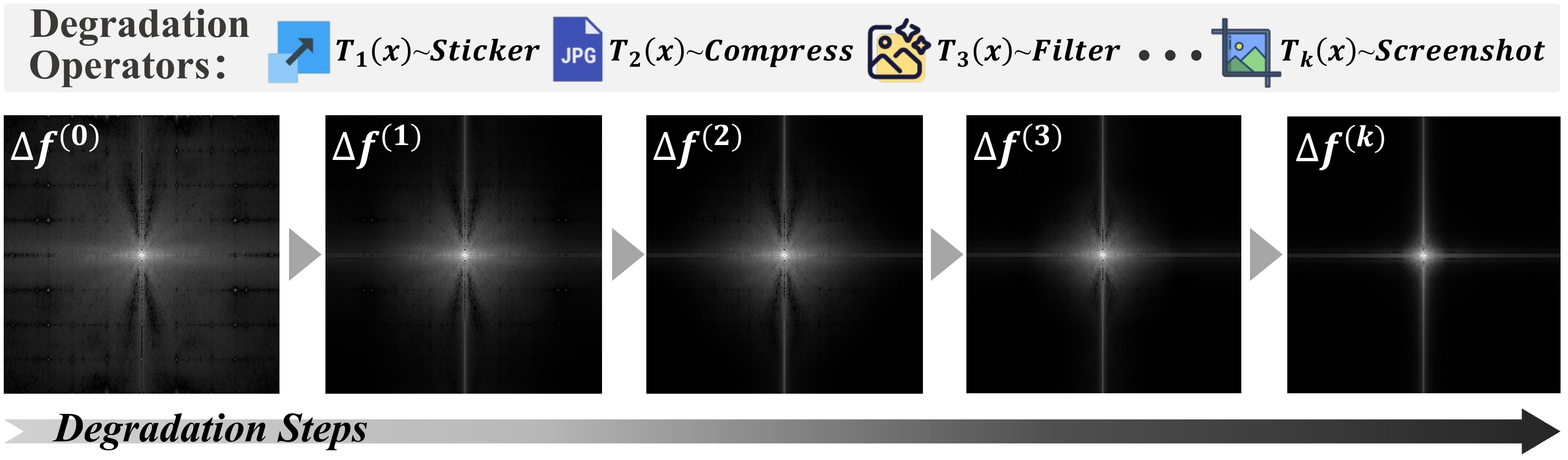}
\caption{
\textbf{Challenges of Chain Degradations.}
As the propagation chain length increases, multi-stage compression and processing progressively suppress generative artifacts, leading to reduced separability between real and fake samples in the frequency domain. 
This explains the sharp performance drop of artifact-driven detectors (e.g., high-frequency artifacts) in real-world conditions.
}
\label{fig_clg1} 

\end{figure}

In practice, the distribution of synthetic images is open and continuously evolving as new architectures and sampling strategies emerge, gradually approaching the distribution of real images (see Fig.~\ref{fig_1}(a) and Fig.~\ref{fig_clg2}).
Therefore, detectors that depend on artifacts from past generators often fail to adapt to future, more realistic synthetic images.
Moreover, nearly all accessible images undergo multiple rounds of propagation and post-processing across different platforms and devices.
As shown in Fig.~\ref{fig_1}(b) and Fig.~\ref{fig_clg1}, this chained propagation and processing further weakens forgery cues, making it even more difficult to distinguish between real and synthetic images

Based on these observations, we argue that the artifact-driven paradigm faces fundamental limitations.
In contrast, the real image distribution is constrained by physical imaging principles and device characteristics, making it inherently stable and learnable \cite{healey2002radiometric,lukas2006digital,brooks2019unprocessing,wang2024adaptiveisp}.
\textbf{To achieve robust detection under real-world conditions, it is crucial to model the real distribution itself, ensuring consistent and stable decision boundaries across quality domains.}

To this end, we propose \textbf{Real-centric Envelope Modeling (REM)}, a new detection paradigm that shifts the focus from learning generator-specific artifacts to modeling the robust distribution of real images.
Specifically, REM first produces diverse near-real samples by applying feature-level perturbations to self-reconstruction models.
These samples preserve semantic consistency with real images while exhibiting slight statistical deviations, enabling the model to learn a compact, smooth envelope around the real distribution.
An Envelope Estimator then explicitly models this boundary in feature space, and a Cross-Domain Consistency constraint ensures its stability under real-world degradations.
Together, these components greatly enhance robustness and generalization to unseen generators.

REM demonstrates outstanding performance across various benchmarks, surpassing state-of-the-art (SOTA) models by an average accuracy of 10.1\% on \textit{AIGCDetect} \cite{zhong2023patchcraft} and by 6.5\% across four in-the-wild datasets (\textit{Chameleon} \cite{yan2024sanity}, etc).
Notably, due to its stable modeling of the real data distribution, REM also shows strong potential in forgery source attribution, achieving SOTA results in both closed-set and open-set settings.

Furthermore, existing detection benchmarks remain confined to idealized settings and lack coverage of real-world propagation scenarios.
To address this limitation, we construct a new dataset named \textbf{\textit{RealChain}}, which collects images generated by the most advanced and widely used commercial and open-source models, and systematically simulates multi-round cross-platform and cross-device propagation as well as user post-processing (chain degradations).
REM still exhibits strong robustness on the \textit{RealChain} benchmark, outperforming SOTA methods by 18.4\%.

Our main contributions are summarized as follows:
\begin{itemize}
    \item We design the Real-centric Envelope Modeling (REM) framework, which relies on real distribution boundary for more reliable discrimination, providing a new perspective for AIGI detection.
    \item We construct the comprehensive \textit{RealChain} dataset covering diverse state-of-the-art generators and chain degradations, enabling evaluation of detection performance under realistic conditions.
    \item Extensive experiments on multiple benchmarks show that REM significantly outperforms existing methods under both degraded and unseen-generator settings, and further demonstrates remarkable competitive potential for forgery source attribution.
\end{itemize}

\section{Related Work}

\noindent\textbf{Artifact-driven AIGI Detection.}  
Early methods for AI-generated image (AIGI) detection rely on identifying generator-specific artifacts.  
CNNSpot \cite{wang2020cnn} first demonstrates that a simple CNN classifier can effectively distinguish synthetic images from known generators but fails to generalize to unseen ones.  
To enhance cross-generator generalization, UnivFD \cite{ojha2023towards} employs a CLIP-based backbone, leveraging large-scale pretraining for better distinguishing forgery types. 
FatFormer \cite{liu2024forgery} and C2P-CLIP \cite{tan2025c2p} further optimize CLIP-based detectors by incorporating textual guidance to refine the representation space.  
NPR \cite{tan2024rethinking} focuses on upsampling artifacts introduced by generative models, while SAFE \cite{li2024improving} exploits frequency inconsistencies as discriminative cues.  
AIDE \cite{yan2024sanity} integrates both frequency and semantic signals to improve detection performance.  

Although artifact-driven detectors achieve high accuracy under controlled settings, they are easily influenced by non-causal low-level cues such as image format and resolution. 
As a result, their performance drops significantly in real-world scenarios where such artifacts are suppressed or when facing more realistic generated images. 
These limitations expose the inherent fragility of the artifact-driven paradigm.

\noindent\textbf{Dataset Alignment AIGI Detection.}  
A recent line of research aims to mitigate evaluation bias by aligning real and generated data. 
FakeInversion \cite{cazenavette2024fakeinversion} introduces a bias-reduced benchmark by matching real and fake samples in both content and style. 
SemGIR \cite{yu2024semgir} and DRCT \cite{chen2024drct} achieve semantic-level alignment through reconstruction or diffusion inversion, while B-Free \cite{guillaro2025bias} and AlignedForensics \cite{rajan2024aligned} use inpainting-based or VAE reconstruction to reduce dataset bias. 
DDA \cite{chen2025dual} further enforces format consistency to prevent shortcut learning. 

These methods show strong generalization improvements, commonly attributed to better semantic alignment and reduced bias. However, we observe that their success stems from capturing the structure of the real image distribution rather than strict pixel-level alignment. In our experiments (see Appendix), deliberately relaxing image correspondence, either by disabling pixel-level alignment or introducing mixed augmentations, not only does not harm performance but can even improve it. \textbf{This suggests that the advantage of dataset alignment may originate from modeling the real distribution boundary through slight perturbations to the image manifold during the alignment process.} Based on this insight, we propose the Real-centric Envelope Modeling (REM) framework, which explicitly models the boundary of the real distribution.

\section{Method}
\subsection{Motivation}
\noindent\textbf{Challenges of Generator Evolution.} The generative landscape itself is open and evolving. 
Existing methods assume that synthetic samples from different generators share common discriminative forgery cues. 
However, as new architectures and sampling strategies continually emerge, the learned artifacts quickly become obsolete, while detectors tend to overfit generator-specific biases. 
Let $\phi(x_r)$ and $\phi(x_f)$ denote feature embeddings of real and synthetic images, respectively. 
For a given generator $g$, the feature discrepancy is defined as:
\begin{equation}
d_g = \| \mathbb{E}_{x_r}[\phi(x_r)] - \mathbb{E}_{x_f \sim g}[\phi(x_f)] \|_2 .
\end{equation}
During training, detectors rely on large $d_g$ values from known generators $\mathcal{G}_{train}$.
However, for an unseen generator $g' \notin \mathcal{G}_{train}$, the discrepancy $d_{g'}$ often decrease,
making its outputs nearly indistinguishable from real images, as illustrated in Fig.~\ref{fig_clg2} (a).

\noindent\textbf{Challenges of Chain Degradations.} Existing artifact-driven detectors are typically designed under high-quality settings, assuming that the input image $x$ has not undergone significant degradation. 
However, in real environments, images are subject to chain degradations, which can be formalized as:
$x^{(k)} = T_k \circ T_{k-1} \circ \cdots \circ T_1(x),$
where $\{T_i\}_{i=1}^k$ represent real-world degradation operators such as compression, filters, digital stickers, etc. 
As the chain length $k$ increases, the low-level statistics of samples are also changed. 
The separability between real and synthetic images can be quantified by their frequency discrepancy:
\begin{equation}
\Delta f^{(k)} = \| F(x_r^{(k)}) - F(x_f^{(k)}) \|_2 ,
\end{equation}
where $F(\cdot)$ denotes the frequency spectrum. 
As $k$ grows, $\Delta f^{(k)}$ gradually decreases, indicating that real and synthetic images become increasingly indistinguishable in the high-frequency domain, as shown in Fig.~\ref{fig_clg1}.
Consequently, artifact-driven detectors show severe performance degradation when exposed to real-world data.

\begin{figure*}[t!]
	\centering
	\includegraphics[scale=0.46]{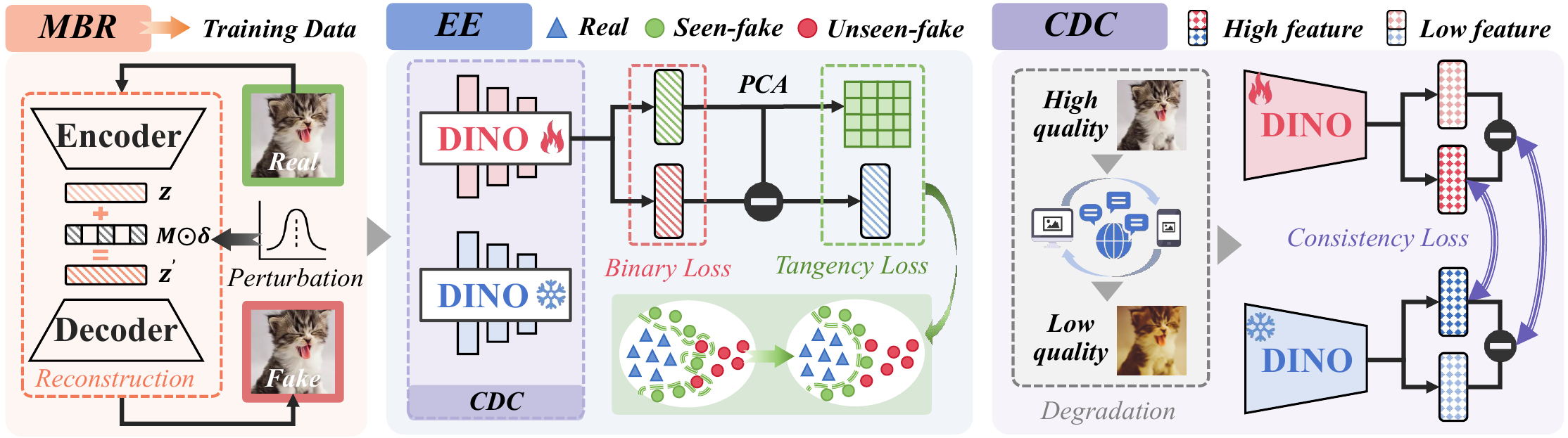}
\caption{
\textbf{Overview of the Real-centric Envelope Modeling (REM) framework.} 
REM consists of three key components: 
(1) Manifold Boundary Reconstruction (MBR) generates diverse near-real samples by applying feature-level perturbations to self-reconstructed inputs; 
(2) Envelope Estimator (EE) then learns a compact and smooth boundary that tightly encloses the real distribution by separating real samples from the MBR generated near-real ones in feature space;
and (3) Cross-domain consistency (CDC) further enforces stability of this boundary under various degradations, ensuring that the learned envelope remains consistent across domains.
}
	\label{fig_ov}
    
\end{figure*}

Artifact-driven detectors have raised concerns about their ability to detect future generators and real-world chain degradations.
To achieve robust detection, as shown in Fig.~\ref{fig_ov}, we propose the \textbf{Real-centric Envelope Modeling (REM)} framework, which anchors detection to the real distribution through three modules: Manifold Boundary Reconstruction for generating near-real samples via feature-level perturbations, Envelope Estimator for boundary learning, and Cross-Domain Consistency for maintaining stability under degradations.

\label{sec:method}

\subsection{Manifold Boundary Reconstruction (MBR)}

The distribution of generated images is open and rapidly evolving, which makes artifact-based detectors trained on limited generators difficult to generalize.  
In contrast, the distribution of real images is governed by physical imaging laws and device constraints, making it relatively stable and learnable.  
Building on this observation, we aim to model the boundary of the real manifold rather than relying on generator-specific artifacts.  
To achieve this, we reconstruct near-real samples around the real distribution and train the model to distinguish them from real samples.  
This process encourages detector to learn a smooth and discriminative envelope that tightly surrounds the real data distribution.

To ensure that the reconstructed samples provide sufficient coverage of the real manifold, we introduce a feature-level perturbation strategy within a variational autoencoder (VAE).  
Specifically, we inject controlled Gaussian noise into randomly selected latent dimensions, allowing the decoder to generate samples that remain visually realistic but deviate slightly from the real manifold in multiple directions.  
Given a real sample $x_r$, its latent representation is obtained as $z = E(x_r)$. 
We then apply random noise to a subset of latent dimensions:
\begin{equation}
z' = z + M \odot \delta, \quad \delta \sim \mathcal{N}(0, \epsilon^2 I),
\end{equation}
where $M$ is a random mask controlling the proportion of perturbed dimensions, and $\epsilon$ controls the magnitude of perturbation, ensuring mild deviations that retain semantic fidelity.
The perturbed vectors are decoded as $x_f = D(z')$, generating semantically consistent yet slightly shifted near-real samples.
By Manifold Boundary Reconstruction, we obtain diverse samples ${x_f}$ forming a local envelope around the real manifold, providing reliable boundary supervision for the Envelope Estimator.

\subsection{Envelope Estimator (EE)} 
With the near-real samples provided by MBR, we explicitly model the boundary of the real data distribution in the feature space. 
The core idea is that real samples should reside inside the distribution envelope, while near-real samples are expected to lie closely along its boundary. 
Formally, the Envelope Estimator aims to learn a smooth and geometrically consistent boundary that encloses the real feature manifold. 

\noindent\textbf{Binary Classification.} 
To coarsely separate real and near-real samples, we adopt a binary cross-entropy loss:
\begin{equation}
\mathcal{L}_{\text{bce}} = 
-\mathbb{E}_{x_r} [\log \sigma(s(h_r))] 
-\mathbb{E}_{x_f} [\log (1 - \sigma(s(h_f)))],
\end{equation}
where $h_r = \phi(x_r)$, $h_f = \phi(x_f)$ denote encoded feature embeddings extracted by encoder $\phi(\cdot)$. 
The function $s(\cdot)$ is a discriminator producing a real-valued score, while $\sigma(\cdot)$ is the sigmoid function converting it into a probability of being real. 
This loss encourages real samples to have higher confidence (i.e., lie inside the envelope) and pushes near-real samples toward the boundary.

\noindent\textbf{Tangency Regularization.}
Classification alone may lead to irregular or overfitted boundaries, especially when near-real samples are unevenly distributed.
To enhance geometric consistency, we introduce a tangency regularization based on the feature displacement between a real sample and its near-real counterpart, defined as $\Delta h = h_f - h_r$.
Let the top-$p$ principal components $U_p$ of real features ${h_r}$ approximate the local tangent space of the real manifold, and construct the projection matrix $P = U_p U_p^\top$.
For each $\Delta h$, the term $(I - P)\Delta h$ represents its deviation orthogonal to the tangent space, representing the component moving away from the real manifold surface.
We therefore minimize
\begin{equation}
\mathcal{L}_{\text{tan}} =
\mathbb{E}_{(h_r, h_f)}
\| (I - P)\Delta h \|_2^2,
\end{equation}
which penalizes off-manifold deviations and encourages near-real features to vary smoothly and consistently along the tangent directions, ultimately yielding a geometry-aligned and continuous envelope boundary.

\subsection{Cross-Domain Consistency (CDC)}

In real-world scenarios, images inevitably undergo multiple stages of compression, noise injection, and re-encoding. 
These operations do not alter semantics but introduce significant feature shifts, causing the decision boundary to become unstable across domains. 
To ensure consistent behavior under such perturbations, we introduce the Cross-Domain Consistency (CDC) module, which enforces stability of the learned envelope boundary.

We adopt DINOv3 \cite{simeoni2025dinov3} as the backbone due to its strong robustness to non-semantic distortions. 
Through large-scale self-supervised pre-training, DINOv3 produces features that are highly invariant to photometric and compression changes while remaining sensitive to semantic content. 
This property makes it an ideal choice for serving as a semantic anchor that preserves high-level meaning under domain degradations. 
In CDC, a frozen DINOv3 acts as the anchor network $f(\cdot)$, providing semantically grounded but non-adaptive representations, while a fine-tuned DINOv3 serves as the learner $\phi(\cdot)$, adapting to the detection task while maintaining alignment with the anchor space.

Given an image $x$ and its degraded $x' = \textit{Degrade}(x)$, we extract anchor features $h = \phi(x)$ and $h' = \phi(x')$, project them through a linear mapping $W$ to obtain $\hat{h}$ and $\hat{h}'$, and compute learner features $a = f(x)$ and $a' = f(x')$. 
Two complementary consistency constraints are applied:

\noindent\textbf{Anchor Consistency.}  
This term keeps the learner’s features close to the anchor space, preventing semantic drift and maintaining the discriminative structure learned from the frozen backbone:
\begin{equation}
\mathcal{L}_{\text{anc}} = \| a - \hat{h} \|_2^2.
\end{equation}
\noindent\textbf{Residual Consistency.}  
This constraint ensures that the residuals between the learner and the anchor remain invariant across domains, effectively preserving the geometric shape of the decision envelope under degradations:
\begin{equation}
\mathcal{L}_{\text{res}} = \| (a - \hat{h}) - (a' - \hat{h}') \|_2^2.
\end{equation}

The low-quality images (\textbf{data augmentation}) in CDC are generated through simulated chain degradations including compression and perturbation (see Appendix). 

\subsection{Overall Objective}

The overall training objective integrates the envelope modeling and cross-domain consistency terms. 
Specifically, the total loss is defined as:
\begin{equation}
\begin{aligned}
\mathcal{L} 
&= \underbrace{ \big( \mathcal{L}_{\text{bce}} + \lambda_1 \mathcal{L}_{\text{tan}} \big) }_{\mathcal{L}_{\text{EE}}}
+ 
\underbrace{ \big( \lambda_2 \mathcal{L}_{\text{anc}} + \lambda_3 \mathcal{L}_{\text{res}} \big) }_{\mathcal{L}_{\text{CDC}}}, 
\end{aligned}
\end{equation}
where the hyperparameters $\lambda_1$, $\lambda_2$, and $\lambda_3$ are empirically determined to balance the importance of each.

\begin{figure}[t]
\centering
\includegraphics[scale=0.158]{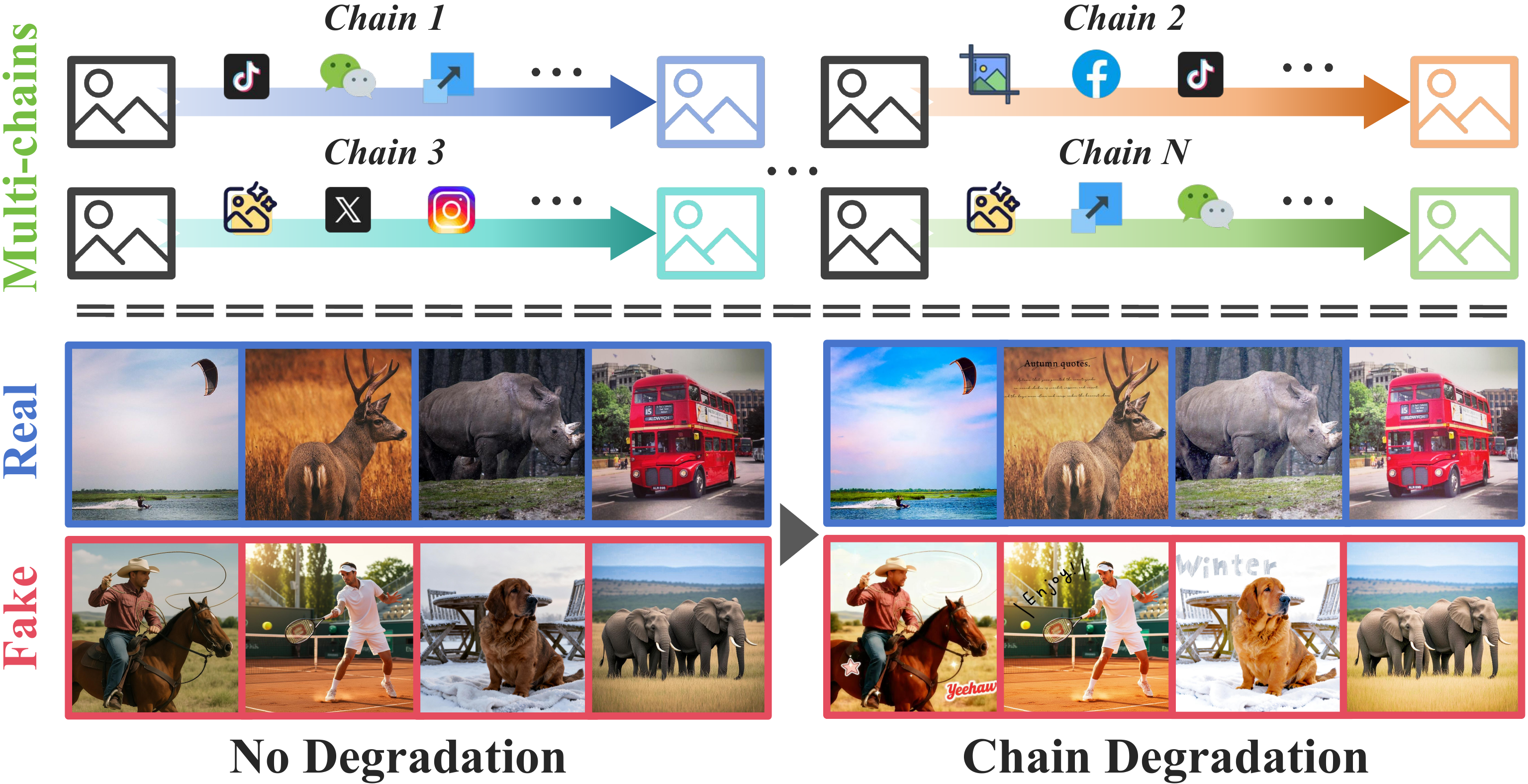}
\caption{
The \textit{RealChain} dataset contains images with no degradation and images with chain degradations.
}
\label{fig_dataset} 

\end{figure}

\section{Dataset Construction}

To evaluate AIGI detection robustness under realistic conditions,
we build \textbf{\textit{RealChain}}, a dataset featuring (i) broad coverage of emerging generative models and (ii) realistic simulation of real-world degradation chains (see Fig.~\ref{fig_dataset}).
\textit{RealChain} reproduces cumulative degradations and distribution shifts in online propagation, providing a comprehensive benchmark for robust AIGI detection.

\subsection{Source of Images}
\noindent\textbf{Fake Images.} To reflect the rapid evolution of generative models, \textit{RealChain} integrates both open-source and commercial sources.  
Open-source models include \textit{QwenImage} \cite{qwenimage_2025}, \textit{SDv3.5} \cite{dm35_2025}, and \textit{Flux.1} \cite{flux_2025}, 
while commercial ones include \textit{Hunyuan 3.0} \cite{hunyuan3_2025}, \textit{NanoBanana} \cite{nanobanana_2025}, and \textit{Seedream 4.0} \cite{seedream4_2025}.  
To capture diverse generation mechanisms, we adopt two primary modes:

\noindent\textbf{(i) Text-to-Image.}
Images are generated from diverse textual prompts to simulate large-scale content creation.
We design 1,000 prompts (500 coarse and 500 fine) and render each using six models, producing 6,000 samples.

\noindent\textbf{(ii) Image-to-Image.}
Given real references, we perform style transfer and local editing to simulate practical post-editing behaviors.
Using 1,000 prompts rendered by \textit{Seedream 4.0} \cite{seedream4_2025}, we obtain 1,000 samples.

\noindent\textbf{Real Images.}
We collect 7,000 real images from MSCOCO \cite{lin2014microsoft}, 
OpenImage-v7 \cite{openimages_web}, Unplash \cite{unsplash_web}, and ImageNet \cite{russakovsky2015imagenet}, ensuring diversity across domains.

\subsection{Chain Degradations Design}

To systematically analyze the effects of real-world propagation and post-processing, we define two dataset settings:

\noindent\textbf{No Degradation (ND).}  
This setting includes directly captured real images and cleanly generated synthetic images. It is used to evaluate detector generalization on the most advanced generators and diverse real sources, serving as the upper performance bound.

\noindent\textbf{Chain Degradations (CD).}  
We reproduce realistic cross-platform and device propagation operations that involve compression, re-encoding, and diverse user post-processing.
Specifically, we define a library of common and easily reproducible degradation operations observed in real-world scenarios:
\textbf{(i) Propagation:} Uploading/downloading via PC applications; uploading/downloading via mobile applications; uploading from PC and downloading on mobile; uploading from mobile and downloading on PC. \textbf{(ii) Post-processing:} Applying filters; inserting stickers; cropping or changing aspect ratio; taking screenshots.
Each sample in the \textit{RealChain} dataset undergoes a random chain of propagation and post-processing operations. Detailed degradation chain configurations are provided in the Appendix.

%表格aigibenchmark-------------------------------
\begin{table*}[t!]
\centering
\caption{Performance comparison on the \textit{AIGCDetect} benchmark. The test benchmark uses balance accuracy (B.Acc). Bold numbers indicate the best performance in each column, and underlined numbers indicate the second-best performance.}
\fontsize{9pt}{11pt}\selectfont   % ====== 调整为9pt ======
\setlength{\tabcolsep}{5pt}
\renewcommand{\arraystretch}{1}
\begin{adjustbox}{max width=\textwidth}
\begin{tabular}{lccccccccccccccccccccc}
\toprule
\textbf{Method} & 
\makecell{ADM} & 
\makecell{DALLE2} & 
\makecell{GLIDE} & 
\makecell{Midj.} & 
\makecell{VQDM} & 
\makecell{Big-\\GAN} & 
\makecell{Cycle-\\GAN} & 
\makecell{Gau-\\GAN} & 
\makecell{Pro-\\GAN} & 
\makecell{SDXL} & 
\makecell{SD1.4} & 
\makecell{SD1.5} & 
\makecell{Star-\\GAN} & 
\makecell{Style-\\GAN} & 
\makecell{Style-\\GAN2} & 
\makecell{WFR} & 
\makecell{Wukong} & 
\textbf{\makecell{Avg\\B.Acc}} &
\textbf{\makecell{Avg\\AP}} \\
\midrule
NPR \cite{tan2024rethinking}               & 43.8 & 20.0 & 41.2 & 53.4 & 48.4 & 53.1 & 76.6 & 42.2 & 58.7 & 59.6 & 55.1 & 55.0 & 67.4 & 57.9 & 54.6 & 58.8 & 57.4 & 53.1 & 46.4\\
UnivFD \cite{ojha2023towards}            & 62.5 & 50.0 & 61.3 & 55.1 & 76.9 & 87.5 & 96.9 & 98.8 & 99.4 & 58.2 & 55.6 & 55.7 & 95.1 & 80.0 & 69.4 & 69.2 & 61.1 & 72.5 & 86.8\\
FatFormer \cite{liu2024forgery}         & 80.2 & 68.5 & 91.1 & 54.4 & 88.0 & 99.2 & 99.5 & 99.1 & 98.5 & 71.7 & 67.5 & 67.2 & 99.4 & 98.0 & 98.8 & 88.3 & 75.6 & 85.0 & 92.0\\
SAFE \cite{li2024improving}               & 49.5 & 49.5 & 53.0 & 49.0 & 50.2 & 52.2 & 51.9 & 50.0 & 50.0 & 49.8 & 49.7 & 49.8 & 50.1 & 50.0 & 50.0 & 49.8 & 50.3 & 50.3 & 52.0\\
C2P-CLIP \cite{tan2025c2p}          & 71.6 & 52.3 & 73.5 & 56.6 & 73.7 & 98.4 & 96.8 & 98.8 & 99.3 & 62.3 & 77.5 & 76.9 & 99.6 & 93.1 & 79.4 & 94.8 & 79.4 & 81.4 & 93.3\\
AIDE \cite{yan2024sanity}              & 52.9 & 51.1 & 60.2 & 49.8 & 69.3 & 70.1 & 93.6 & 60.6 & 89.0 & 49.6 & 51.6 & 51.0 & 72.1 & 66.5 & 59.0 & 80.6 & 54.5 & 63.6 & 79.2\\
DRCT \cite{chen2024drct}              & 79.9 & 89.2 & 89.2 & 85.5 & 88.6 & 81.4 & 91.0 & 93.8 & 71.1 & 88.3 & 91.4 & 91.0 & 53.0 & 62.7 & 63.8 & 73.9 & 90.8 & 81.4 & 90.7\\
Aligned \cite{rajan2024aligned}      & 51.6 & 52.0 & 55.6 & 96.2 & 72.1 & 51.2 & 49.5 & 50.8 & 50.7 & 95.1 & 99.7 & 99.6 & 53.8 & 52.7 & 51.6 & 50.0 & 99.6 & 66.6 & 80.7\\
DDA \cite{chen2025dual}     & 89.5 & 94.6 & 89.6 & 95.6 & 76.6 & 91.0 & 72.5 & 92.7 & 92.8 & 99.4 & 98.7 & 98.6 & 72.7 & 87.8 & 90.2 & 52.1 & 98.8 & \underline{87.8} &\underline{98.9} \\
\midrule
\rowcolor{blue!5}
\textbf{REM}     & 98.1 & 99.3 & 98.6 & 96.5 & 99.3 & 99.4 & 98.1 & 99.8 & 97.4 & 99.8 & 99.3 & 99.5 & 90.2 & 95.1 &  96.3 &98.4 & 99.5 & \textbf{97.9} & \textbf{99.9}\\

\bottomrule
\end{tabular}
\end{adjustbox}

\label{tab:aigc}
\end{table*}

% 表格 REM benchmark ------------------------------------
\begin{table*}[t!]
\centering
\caption{Performance comparison on the \textit{DRCT-2M} benchmark.}

\fontsize{9pt}{11pt}\selectfont
\setlength{\tabcolsep}{5pt}
\renewcommand{\arraystretch}{1}
\begin{adjustbox}{max width=\textwidth}
\begin{tabular}{lcccccccccccccccccccc}
\toprule
\textbf{Method} &
LDM &
SDv1.4 &
SDv1.5 &
SDv2 &
SDXL &
\makecell{SDXL-\\Refiner} &
\makecell{SD-\\Turbo} &
\makecell{SDXL-\\Turbo} &
\makecell{LCM-\\SDv1.5} &
\makecell{LCM-\\SDXL} &
\makecell{SDv1-\\Ctrl} &
\makecell{SDv2-\\Ctrl} &
\makecell{SDXL-\\Ctrl} &
\makecell{SDv1-\\DR} &
\makecell{SDv2-\\DR} &
\makecell{SDXL-\\DR} &
\textbf{\makecell{Avg\\B.Acc}} &
\textbf{\makecell{Avg\\AP}} \\
\midrule
NPR \cite{tan2024rethinking} & 33.0 & 29.1 & 29.0 & 35.1 & 33.2 & 28.4 & 27.9 & 27.9 & 29.4 & 30.2 & 28.4 & 28.3 & 34.7 & 67.9 & 67.4 & 66.1 & 37.3 & 40.3 \\ 
UnivFD \cite{ojha2023towards} & 85.4 & 56.8 & 56.4 & 58.2 & 63.2 & 55.0 & 56.5 & 53.0 & 54.5 & 65.9 & 68.0 & 65.4 & 75.9 & 64.6 & 56.2 & 53.9 & 61.9 & 85.7\\ 
FatFormer \cite{liu2024forgery} & 55.9 & 48.2 & 48.2 & 48.2 & 48.2 & 48.3 & 48.2 & 48.2 & 48.3 & 50.6 & 49.7 & 49.9 & 59.8 & 66.3 & 60.6 & 56.0 & 49.4 & 47.8\\ 
SAFE \cite{li2024improving} & 50.3 & 50.1 & 50.0 & 50.0 & 49.9 & 50.1 & 50.0 & 50.0 & 50.1 & 50.0 & 49.9 & 50.0 & 54.7 & 98.2 & 98.5 & 97.3 & 62.4 & 57.7\\ 
C2P-CLIP \cite{tan2025c2p} & 83.0 & 51.7 & 51.7 & 52.9 & 51.9 & 64.6 & 51.7 & 50.6 & 52.0 & 66.1 & 56.9 & 54.7 & 77.8 & 67.2 & 57.1 & 56.7 & 65.4 & 70.7 \\ 
AIDE \cite{yan2024sanity} & 64.4 & 74.9 & 75.1 & 58.5 & 53.5 & 66.3 & 52.8 & 52.8 & 70.0 & 54.3 & 65.9 & 53.6 & 53.9 & 95.3 & 73.3 & 69.0 & 73.7 & 70.2\\ 
DRCT \cite{chen2024drct} & 96.7 & 96.3 & 96.3 & 94.9 & 96.2 & 93.5 & 93.4 & 92.9 & 91.2 & 95.0 & 95.6 & 92.7 & 92.0 & 94.1 & 69.6 & 57.4 & 92.4 & 96.1\\ 
Aligned \cite{rajan2024aligned} & 99.9 & 99.9 & 99.9 & 99.6 & 90.2 & 81.3 & 99.7 & 89.4 & 99.7 & 90.0 & 99.9 & 99.2 & 87.6 & 99.9 & 99.8 & 92.6 & 95.5 & 99.8\\ 
DDA \cite{chen2025dual} & 99.2 & 98.9 & 99.0 & 98.3 & 98.0 & 96.8 & 97.9 & 94.8 & 95.9 & 98.2 & 98.7 & 99.0 & 99.4 & 99.0 & 99.5 & 96.3 & \underline{98.1} & \underline{99.9} \\ 
\midrule
\rowcolor{blue!5}
\textbf{REM} & 100.0 & 99.6 & 99.5 & 97.6 & 99.1 & 97.0 & 99.7 & 99.0 & 99.1 & 99.5 & 100.0 & 99.9 & 100.0 & 98.2 & 99.2 & 95.2 & \textbf{98.9} & \textbf{100.0} \\
\bottomrule
\end{tabular}
\end{adjustbox}

\label{tab:drct}
\end{table*}

%表格DDACOCO和wild并列（左右互换）----------------------------------------
\begin{table*}[!t]
\setlength{\tabcolsep}{5pt}
\renewcommand{\arraystretch}{1}
\centering  % �� 添加这个，确保总宽度居中分配
% ----------- 左侧：Wild Setting -----------
\begin{minipage}[t]{0.637\linewidth}
\raggedright  % 左对齐，使表靠左页边
\caption{Comparison on the \textit{Chameleon}, \textit{SynthWildx}, \textit{WildRF}, and \textit{AIGIBench}.}
\begin{adjustbox}{width=\linewidth}
\begin{tabular}{l c | c c c | c c c | cc|cc}
\toprule
\multirow{2}{*}{\textbf{Method}} & \multirow{2}{*}{\textbf{\makecell{Cham-\\eleon}}} &
\multicolumn{3}{c|}{\textbf{SynthWildx}} &
\multicolumn{3}{c|}{\textbf{WildRF}} &
\multicolumn{2}{c|}{\textbf{AIGIBench}} &
\multirow{2}{*}{\textbf{\makecell{Avg\\B.Acc}}} &
\multirow{2}{*}{\textbf{\makecell{Avg\\AP}}} 
\\

\cmidrule(lr){3-5} \cmidrule(lr){6-8} \cmidrule(lr){9-10}
 &  & DALLE3 & Firefly & Midj.
 & Facebook & Reddit & Twitter & SocRF  & ComAI  \\
\midrule
NPR \cite{tan2024rethinking}              & 59.9 & 43.6 & 61.3 & 44.5 & 78.1 & 61.0 & 51.3 & 53.3 & 55.1 & 57.1 & 57.6\\
UnivFD \cite{ojha2023towards}           & 50.7 & 45.4 & 65.3 & 46.2 & 49.1 & 60.2 & 56.5 & 54.6 & 51.4 & 53.3 & 52.5\\
FatFormer \cite{liu2024forgery}        & 51.2 & 46.5 & 61.6 & 48.3 & 54.1 & 68.1 & 54.4 & 56.9 & 51.9 & 54.9 & 64.0\\
SAFE \cite{li2024improving}             & 59.2 & 49.4 & 48.2 & 49.6 & 50.9 & 74.1 & 37.5 & 58.4 & 54.5 & 52.7 & 55.4\\
C2P-CLIP \cite{tan2025c2p}         & 51.1 & 56.9 & 61.4 & 53.0 & 54.4 & 68.4 & 55.9 & 56.4 & 51.0 & 57.3 & 58.8\\
AIDE \cite{yan2024sanity}             & 63.1 & 63.4 & 48.8 & 51.9 & 57.8 & 71.5 & 45.8 & 57.3 & 53.7 & 57.5 & 51.0\\
DRCT \cite{chen2024drct}             & 56.6 & 58.3 & 56.4 & 50.5 & 46.6 & 53.1 & 55.2 & 67.5 & 76.7 & 53.8& 62.4\\
Aligned \cite{rajan2024aligned} & 71.0 & 85.5 & 58.5 & 92.2 & 89.4 & 69.1 & 81.8 & 73.5 & 71.9 & 78.2 & 86.1\\
DDA \cite{chen2025dual}              & 82.4 & 92.3 & 87.3 & 93.1 & 93.1 & 86.4 & 91.5 & 79.6 & 88.6 & \underline{88.3}& \underline{93.4}\\
\midrule
\rowcolor{blue!5}
\textbf{REM} & 91.3 & 96.5 & 92.8 & 96.9 &  97.2 & 98.1 & 98.2 & 92.2 & 90.4& \textbf{94.8}& \textbf{97.4}\\
\bottomrule
\end{tabular}
\end{adjustbox}
\label{tab:wild}
\end{minipage}%
\hfill
% ----------- 右侧：No-Degradation / Chain-Degradation -----------
\begin{minipage}[t]{0.34\linewidth}
\raggedleft  % 右对齐，使表贴右页边
\caption{Comparison on the \textit{RealChain}.}
\begin{adjustbox}{width=\linewidth}
\begin{tabular}{l|ccc|ccc}
\toprule
\multirow{2}{*}{\textbf{Method}} &
\multicolumn{3}{c|}{\textbf{No Degradation}} &
\multicolumn{3}{c}{\textbf{Chain Degradations}} \\
\cmidrule(lr){2-4} \cmidrule(lr){5-7}
& R.Acc & F.Acc & B.Acc
& R.Acc & F.Acc & B.Acc \\
\midrule
NPR               & 53.1 & 95.5 & 74.3 & 73.5 & 37.9 & 55.7 \\
UnivFD           & 96.8 & 11.6 & 54.2  & 95.8 & 6.9 & 51.3\\
FatFormer         & 95.7 & 14.0 & 54.9 & 98.3 & 4.1 & 51.2\\
SAFE               & 98.2 & 24.5 & 61.4 & 99.3 & 0.3 & 49.8\\
C2P-CLIP          & 98.3 & 4.2 & 51.3 & 98.3 & 4.3 & 51.3\\
AIDE            & 97.2 & 23.6 & 60.4 & 98.8 & 1.3 & 50.0\\
DRCT           & 95.9 & 42.3 & 69.1 & 92.0 & 18.9 & 55.4\\
Aligned        & 99.5 & 37.9 & 68.7 & 99.8 & 16.4 & 58.0\\
DDA           & 89.5 & 88.2 & \underline{88.8} & 79.3 & 52.4 & \underline{65.8}\\
\midrule
\rowcolor{blue!5}
\textbf{REM} & 94.7 & 87.9 & \textbf{91.3} & 85.3 & 83.0 & \textbf{84.2} \\
\bottomrule
\end{tabular}
\label{tab:chain}
\end{adjustbox}
\end{minipage}

\end{table*}

% ==== 表2：Backbone Comparison (single-column) ====
\begin{table}[t]
\centering
\setlength{\tabcolsep}{3pt}
\renewcommand{\arraystretch}{1.05}
\caption{Performance of Advanced Baselines.}

\fontsize{8.5pt}{10pt}\selectfont
\begin{adjustbox}{width=0.47\textwidth}
\begin{tabular}{l|cccccc|cc|c}
\toprule
\multirow{2}{*}{\makecell{\textbf{Method}\\ (Data Augmentation)}} & 
\multirow{2}{*}{\makecell{\textit{AIGI}-\\\textit{Detect}}} &
\multirow{2}{*}{\makecell{\textit{DRCT}-\\\textit{2M}}} &
\multirow{2}{*}{\makecell{\textit{Cham}-\\\textit{eleon}}} &
\multirow{2}{*}{\makecell{\textit{Synth}-\\\textit{Wildx}}} &
\multirow{2}{*}{\makecell{\textit{WildRF}}} &
\multirow{2}{*}{\makecell{\textit{AIGI}-\\\textit{Bench}}} &
\multicolumn{2}{|c|}{\textit{RealChain}} &
\multirow{2}{*}{\textbf{Avg}} \\
\cmidrule(lr){8-9}
 & & & & & & & \textit{ND} & \textit{CD} & \\
\midrule
AIDE+Genimage & 80.2 & 79.1 & 68.3 & 63.3 & 65.6 & 60.1 & 67.5 & 59.8 & 68.0  \\
AIDE+DDA & 87.7 & 88.4 & 73.3 & 63.9 & 65.4 & 65.2 & 74.3 & 63.5 & 72.7  \\
DINOv3+Genimage  & 92.1 & 98.0 & 85.9 & 88.5 & 89.2 & 85.8 & 86.2 & 73.4 & 87.4  \\
DINOv3+DDA & 94.5 & 97.8 & 87.5 & 90.7 & 95.4 & 87.8 & 91.0 & 77.3 & 90.3  \\
\midrule
\rowcolor{blue!5}
DINOv3+REM & \textbf{97.9} & \textbf{98.9} & \textbf{91.3} & \textbf{95.4} & \textbf{97.9} & \textbf{91.3} & \textbf{91.3} & \textbf{84.2} & \textbf{93.5} \\
\bottomrule
\end{tabular}
\end{adjustbox}
\label{tab:baseline}
\end{table}

\section{Experiments}

We conduct extensive experiments to validate the effectiveness and robustness of our proposed REM under both ideal and real-world conditions. 

\subsection{Experimental Settings}

\noindent\textbf{Datasets.}  
\textbf{(1) Ideal benchmark.}  
We first evaluate REM on two clean and high-quality datasets, \textit{AIGCDetect} \cite{zhong2023patchcraft} and \textit{DRCT-2M} \cite{chen2024drct}.  
\textit{AIGCDetect} \cite{zhong2023patchcraft} contains 17 types of generator data, including GAN-based and Diffusion-based models, while \textit{DRCT-2M} \cite{chen2024drct} includes 16 generators with diverse architectures and sampling strategies.  
These datasets serve as controlled references for assessing the upper bound of generalization performance under ideal conditions.
\textbf{(2) Real-world degradation benchmark.}  
We evaluate REM on both public and newly collected real-world datasets, including \textit{Chameleon} \cite{yan2024sanity}, \textit{SynthWildx} \cite{cozzolino2024raising}, \textit{WildRF} \cite{cavia2024real}, \textit{AIGIBench} \cite{li2025artificial}, and our \textit{RealChain} dataset. 
These datasets cover diverse propagation environments, black-box post-processing, and multiple image formats generated by both open-source and commercial models. 
They collectively reflect the complexity of real social media scenarios and serve to assess the model’s robustness and generalization under uncontrolled degradations.
% \textbf{(3) Simulated extreme degradation benchmark.}  
% To emulate future scenarios with more severe and complex degradations, we apply a simulated degradation chain to the \textit{RealChain} dataset, defined as $x^{(k)} = T_k \circ \cdots \circ T_1(x)$, where $T(x)$ denotes a degradation operation and $k$ represents the degradation depth.  
% This setting is designed to model the cumulative effects of sequential degradations and to evaluate how detection performance evolves as $k$ increases.

\begin{figure*}[t!]
	\centering
	\includegraphics[scale=0.432]{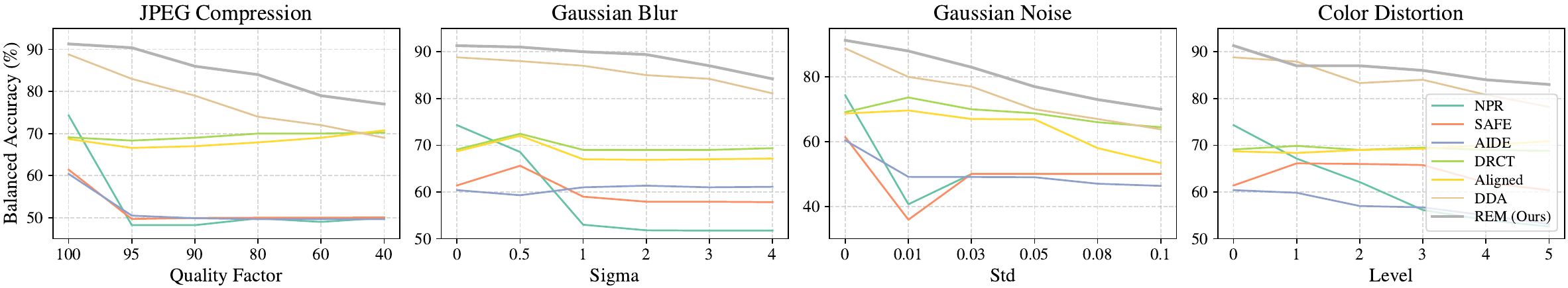}
    
    \caption{
    REM is evaluated under various perturbations, including JPEG compression, Gaussian blur, Gaussian noise, and color distortion. 
}
	\label{fig_rob}
    
\end{figure*}

\noindent\textbf{Evaluation Protocol.}
We evaluate performance with four common metrics: Real Accuracy (\textit{R.Acc}), Fake Accuracy (\textit{F.Acc}), Balanced Accuracy (\textit{B.Acc}), and Average Precision (\textit{AP}).
These metrics provide a more reliable and balanced assessment of model performance.
We compare REM with a series of existing methods, including artifact-driven detectors \cite{tan2024rethinking, li2024improving, ojha2023towards, liu2024forgery, tan2025c2p, yan2024sanity}, as well as alignment-based approaches \cite{chen2024drct, rajan2024aligned, chen2025dual}.
Following the experimental settings of DDA \cite{chen2025dual}, all baseline results are obtained using the official implementations and pretrained weights released by their authors, with JPEG quality alignment applied for the \textit{AIGCDetect} \cite{zhong2023patchcraft} . To further highlight our REM paradigm, we also evaluate the advanced artifact-driven method AIDE \cite{yan2024sanity} and visual backbone DINOv3 \cite{simeoni2025dinov3}, which are trained respectively on a more diverse generative image dataset (GenImage \cite{zhu2023genimage}) and on the DDA dataset \cite{chen2025dual} (both use data augmentation of CDC module).

\noindent\textbf{Implementation Details}
In all comparative and ablation experiments, we adopt DINOv3 ViT-H+/16 \cite{simeoni2025dinov3} as the backbone network and fine-tune it using the LoRA \cite{hu2022lora} algorithm.  
In the preprocessing stage, images with resolutions larger than $512 \times 512$ are resized proportionally so that the shorter side is 512. During training, we use randomly cropped windows of $224 \times 224$, while during inference we use center-cropped windows of $224 \times 224$.
MSCOCO \cite{lin2014microsoft} samples are used as the real-image sources, and near-real samples are constructed via the MBR module.  
We employ the Adam \cite{kingma2014adam} optimizer with a learning rate of 0.0001 and a batch size of 256.  
Training is performed for only one epoch on four NVIDIA A100 GPUs, taking only 30 minutes to complete.

\begin{figure}[t!]
\centering
\includegraphics[scale=0.46]{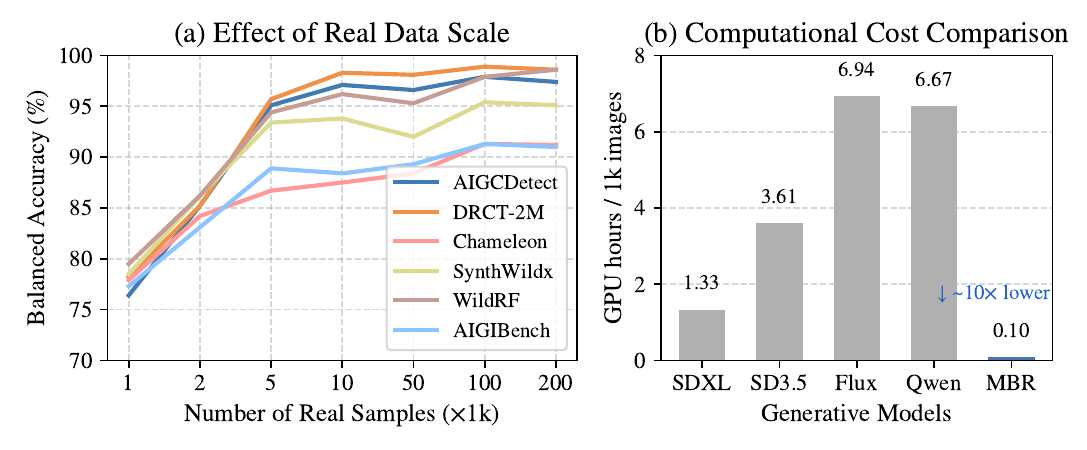}

\caption{
(a) Effect of real data scale. Accuracy and stability improve with more real samples, reaching optimal performance at 100k.
(b) Computational cost. The MBR module requires over \textbf{10$\times$ lower} GPU cost than diffusion-based generators.
}
\label{fig_time} 

\end{figure}

\subsection{Main Results}
\noindent\textbf{Performance on Ideal Benchmarks.}  
As shown in Table~\ref{tab:aigc} and Table~\ref{tab:drct}, REM achieves state-of-the-art results on \textit{AIGCDetect} \cite{zhong2023patchcraft} and \textit{DRCT-2M} \cite{chen2024drct}. 
It consistently outperforms artifact-driven methods such as SAFE \cite{li2024improving} and AIDE \cite{yan2024sanity}, as well as alignment-based approaches including Aligned \cite{rajan2024aligned} and DDA \cite{chen2025dual}. 
REM achieves an average \textit{B.Acc} of 97.9\% and 98.9\% on the two datasets, respectively, outperforming the strongest baselines by 10.1\% and 0.8\%.
These results demonstrate that modeling the real-distribution envelope leads to more generator-agnostic representations than learning generator-specific artifacts.

\noindent\textbf{Performance under Real-world Degradations.}  
As shown in Table~\ref{tab:wild}, we first evaluate REM on four public in-the-wild benchmarks, \textit{Chameleon} \cite{yan2024sanity}, \textit{SynthWildx} \cite{cozzolino2024raising}, \textit{WildRF} \cite{cavia2024real}, and \textit{AIGIBench} \cite{li2025artificial}, which contain naturally propagated images with diverse compression and post-processing operations. 
Across all four datasets, REM achieves the best overall performance, surpassing DDA \cite{chen2025dual} by 6.5\% in average \textit{B.Acc}, and showing a remarkable 8.9\% improvement over DDA on \textit{Chameleon}.  
We further evaluate REM on our newly collected \textit{RealChain} (Table~\ref{tab:chain}), which captures the full complexity of real social media environments. 
REM demonstrates exceptional robustness, maintaining a \textit{B.Acc} of 84.2\% , while all baseline methods almost fail, with only DDA retaining a \textit{B.Acc} above 60\%.
These results confirm that REM effectively handles common degradations and distribution shifts in real-world networks.

\noindent\textbf{Performance of Advanced Baselines.}
To further highlight the advantages of the REM paradigm, we evaluate two advanced baselines AIDE and DINOv3, which are trained respectively on the large-scale and diverse GenImage \cite{zhu2023genimage} dataset and the DDA-aligned dataset \cite{chen2025dual}.
As shown in Table~\ref{tab:baseline}, both methods achieve noticeably better performance when trained on the DDA-aligned dataset than on GenImage, indicating that even imperfect real distribution modeling methods are superior to large-scale artifact-based learning paradigms.
By explicitly modeling the real-distribution envelope, REM captures more robust and generator-agnostic representations than artifact-driven or alignment-based approaches, achieving the best performance across nearly all benchmarks.

\subsection{Analysis and Discussion}

% ==== 表1：Forgery Source Attribution (single-column) ====
\begin{table}[t!]
\centering
\setlength{\tabcolsep}{4pt}
\renewcommand{\arraystretch}{1.05}
\caption{Forgery source attribution on the \textit{AIGCDetect}.}

\fontsize{8.5pt}{10pt}\selectfont
\begin{adjustbox}{width=0.46\textwidth}
\begin{tabular}{l|ccc|ccc|c}
\toprule
\multirow{2}{*}{\textbf{Method}} &
\multicolumn{3}{c|}{\textbf{Closed-set}} &
\multicolumn{3}{c|}{\textbf{Open-set}} &
\multirow{2}{*}{\textbf{Avg}} \\
\cmidrule(lr){2-4} \cmidrule(lr){5-7}
& 4-class & 8-class & 16-class & 4-class & 8-class & 16-class & \\
\midrule
% NPR\cite{tan2024rethinking}     & 78.2 & 65.4 & 52.3 & 50.8 & 37.2 & 26.1 & 51.7 \\
UnivFD \cite{ojha2023towards}   & 82.3 & 71.8 & 58.7 & 23.6 & 15.5 & 11.8 & 43.9  \\
AIDE \cite{yan2024sanity}       & 85.6 & 75.2 & 63.4 & 27.8 & 18.3 & 15.5 & 47.6 \\
Aligned \cite{rajan2024aligned} & 91.8 & 82.7 & 70.2 & 53.5 & 46.1 & 31.7 & 62.7 \\
DDA \cite{chen2025dual}         & 92.5 & 85.3 & 74.1 & 61.2 & 52.7 & 42.0 & 68.0 \\
\midrule
\rowcolor{blue!5}
\textbf{REM} & \textbf{94.6} & \textbf{89.3} & \textbf{80.5} & \textbf{72.1} & \textbf{64.5} & \textbf{53.3} & \textbf{75.7} \\
\bottomrule
\end{tabular}
\end{adjustbox}
\label{tab:attribution}

\end{table}

\noindent\textbf{Robustness to Perturbations.}
We further evaluate REM on the \textit{RealChain} (ND) under various perturbations, including JPEG compression, Gaussian blur, Gaussian noise addition, and color distortion.
As shown in Fig.~\ref{fig_rob}, artifact-based baselines perform poorly under nearly all perturbations. Both our REM and DDA \cite{chen2025dual}
achieve strong results on clean data, while REM remains more stable under perturbations, further demonstrating the advantage of its cross-quality consistency constraint. Interestingly, the alignment-based methods Aligned \cite{rajan2024aligned} and DRCT \cite{chen2024drct} maintain remarkably steady performance across all perturbations. We speculate that this robustness arises because these methods do not explicitly model format or quality factors during training, which causes them to lose discriminative cues that are beneficial only under high-quality conditions. Their weaker performance on clean data further supports this explanation.

\noindent\textbf{Effect of Real Data Scale.}  
Since REM explicitly models the boundary of the real distribution, we analyze how the scale of real data affects its performance. 
As shown in Fig.\ref{fig_time} (a), both accuracy and stability consistently improve as the amount of real samples increases. 
When the number of real samples reaches 100k, REM achieves optimal overall performance across all benchmarks. 
This trend indicates that a richer real distribution allows the model to learn a more complete and robust envelope boundary. 
Fortunately, collecting real data is inexpensive, and the computational cost of the MBR module for reconstructing near-real samples is much lower than that of generating diverse fake images from large diffusion models (see Fig.\ref{fig_time} (b)).

\noindent\textbf{Forgery Source Attribution.}
Fig.~\ref{fig_clg2} shows the feature embeddings learned by REM, where samples from different generators form distinct clusters.
This indicates that, beyond distinguishing real from fake images, REM also captures distributional differences among generators, acquiring forgery source attribution ability without explicit supervision.
We attribute this to REM’s real-centric modeling, which anchors learning on the real distribution and highlights generator-specific variations.
As shown in Table~\ref{tab:attribution}, we evaluate both closed-set and open-set attribution (see Appendix for details).
In the closed-set setting, a classification head is retrained for each method, while in the open-set setting, attribution relies on feature distance.
REM achieved the best results in all settings, with an average accuracy exceeding DDA \cite{chen2025dual} by 7.7\%, confirming its strong representation and robust attribution capabilities.

\begin{figure}[t!]
\centering
\includegraphics[scale=0.17]{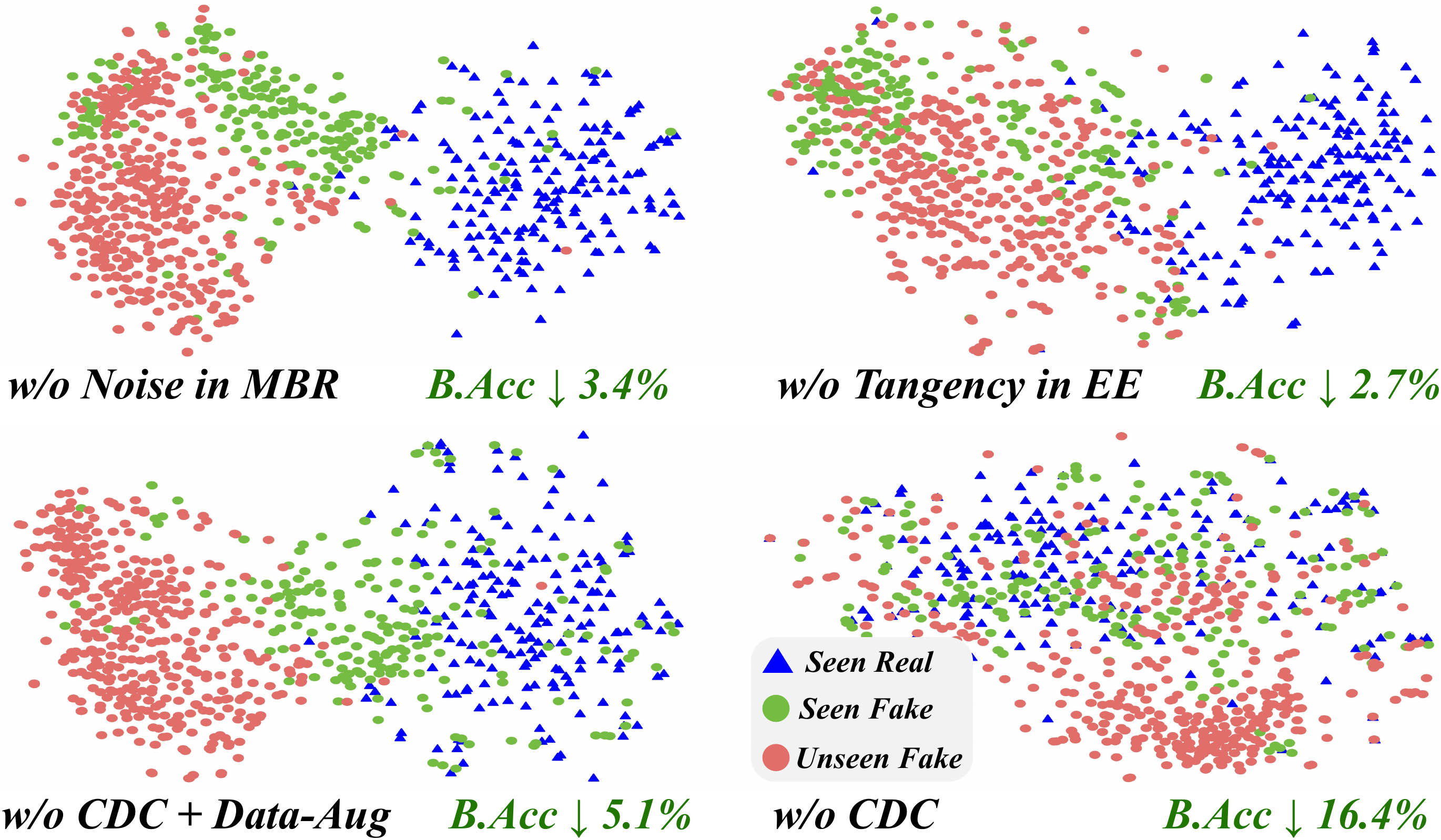}
    \caption{
\textbf{Ablation results of REM.} Each component is evaluated under chain degradations setting on \textit{RealChain} (CD).
    } 
\label{fig_ab} 

\end{figure}

\subsection{Ablation Studies}

We conduct a comprehensive ablation study to analyze the contribution of each component in REM (see Fig.~\ref{fig_ab}).

\noindent\textbf{Manifold Boundary Reconstruction (MBR).}
Replacing the proposed feature-level perturbation with plain VAE reconstruction leads to a 3.4\% drop in \textit{B.Acc} under chain degradations.
Without MBR, near-real samples lack diversity and fail to cover the real boundary.
% In contrast, feature-level perturbations introduce controlled deviations while preserving semantics, enriching manifold coverage.

\noindent\textbf{Envelope Estimator (EE)}
When the tangency loss is removed, the model relies solely on the binary classification loss, making the detector more sensitive to image quality degradation, with the balanced accuracy dropping by 2.7\%.
Feature visualizations further show that the decision surface becomes overfitted and biased toward unseen fake clusters.
% In contrast, incorporating the tangency constraint enforces local alignment of near-real samples along the real manifold, resulting in a smoother and more stable envelope.

\noindent\textbf{Cross-Domain Consistency (CDC)}
Removing CDC leads to a sharp performance drop under chain degradations (\textit{B.Acc} decreases by 16.4\%).
Although data augmentation during training can partially mitigate the impact of degradations, it often results in unstable convergence (see Fig.~\ref{fig_ab}).
% In contrast, CDC explicitly aligns the feature residuals between degraded and clean versions of the same image, thereby maintaining a consistent decision boundary.

% ==== 表2：Backbone Comparison (single-column) ====
\begin{table}[t!]
\centering
\setlength{\tabcolsep}{3pt}
\renewcommand{\arraystretch}{1.05}
\caption{Backbone comparison across datasets.}

\fontsize{8.5pt}{10pt}\selectfont
\begin{adjustbox}{width=0.47\textwidth}
\begin{tabular}{l|cccccc|cc|c}
\toprule
\multirow{2}{*}{\makecell{\textbf{Backbone}\\(DINO)}} & 
\multirow{2}{*}{\makecell{\textit{AIGI}-\\\textit{Detect}}} &
\multirow{2}{*}{\makecell{\textit{DRCT}-\\\textit{2M}}} &
\multirow{2}{*}{\makecell{\textit{Cham}-\\\textit{eleon}}} &
\multirow{2}{*}{\makecell{\textit{Synth}-\\\textit{Wildx}}} &
\multirow{2}{*}{\makecell{\textit{WildRF}}} &
\multirow{2}{*}{\makecell{\textit{AIGI}-\\\textit{Bench}}} &
\multicolumn{2}{|c|}{\textit{RealChain}} &
\multirow{2}{*}{\textbf{Avg}} \\
\cmidrule(lr){8-9}
 & & & & & & & \textit{ND} & \textit{CD} & \\
\midrule
v2-L/14 & 95.1 & 97.9 & 84.6 & 91.4 & 95.8 & 87.2 & 89.3  & 78.5 & 90.0 \\
v2-g/14 & 96.3 & 97.8 & 85.5 & 91.0 & 95.8 & 89.8 & 89.4 & 80.7 & 90.8  \\
v3-L/16 & 97.0 & 98.1 & 89.7 & 93.2 & 96.3 & 89.1 & 91.1  & 83.3 & 92.2 \\
v3-H+/16  & 97.9 & 98.9 & 91.3 & 95.4 & 97.9 & 91.3 & 91.3  & 84.2 & 93.5 \\
v3-7B/16 & \textbf{98.2} & \textbf{99.4} & \textbf{93.7} & \textbf{96.5} & \textbf{98.2} & \textbf{93.7} & \textbf{93.6} & \textbf{86.1} & \textbf{94.9} \\
\bottomrule
\end{tabular}
\end{adjustbox}
\label{tab:dino_backbones}

\end{table}

\noindent\textbf{Influence of Different Backbones.}
We further compare REM’s performance across different DINO backbones.
As shown in Table~\ref{tab:dino_backbones}, REM maintains consistently strong performance across all backbones and improves further with increased pretraining quality.
This demonstrates the robustness of REM’s real-distribution modeling and its ability to efficiently activate and leverage the pretrained knowledge of DINO-based architectures.

\noindent\textbf{How REM Learns Robust Real Distribution.}
MBR introduces controllable feature perturbations, generating near-real samples that expand the neighborhood around the real manifold.
EE draws a discriminative boundary between real and near-real samples, effectively approximating the envelope of the real distribution.
CDC constraint further ensures that representations of the same image remain consistent across different quality levels, keeping the boundary stable under various degradations.
Together, these mechanisms enable REM to capture a compact, smooth, and domain-invariant real-centric envelope, resulting in strong robustness to unseen generators and real-world degradations.

\section{Conclusion}
We introduced REM, a real-centric modeling for detecting AI-generated images. REM achieves robust and generalizable performance under real-world degradation conditions, while enabling effective forgery source attribution.
Our proposed \textit{RealChain} with emerging generators and chain degradations supports real-world evaluation.
Future work will extend REM to video forensics and explore more fine-grained modeling of real distributions.

% --- 7. 参考文献 (改为编号形式) ---
{
    \small
    % unsrt: 按引用顺序编号 [1], [2]... 
    % 如果想按字母顺序编号，请改为 plain
    \bibliographystyle{unsrt} 
    \bibliography{main}
}

% --- 8. 附录 ---
% \clearpage
% \appendix
% \input{sec/X_suppl}

\end{document}